\documentclass{article}

\usepackage{microtype}
\usepackage{graphicx}
\usepackage{subcaption}
\usepackage{booktabs} 

\usepackage{hyperref}

\usepackage[accepted]{icml2026}

\usepackage{amsmath}
\usepackage{amssymb}
\usepackage{mathtools}
\usepackage{amsthm}

\usepackage[capitalize,noabbrev]{cleveref}

\theoremstyle{plain}

\theoremstyle{definition}

\theoremstyle{remark}

\usepackage[textsize=tiny]{todonotes}

\usepackage{amsmath,amsfonts,bm}

\def\eqref#1{equation~\ref{#1}}

\def\1{\bm{1}}

\DeclareMathAlphabet{\mathsfit}{\encodingdefault}{\sfdefault}{m}{sl}
\SetMathAlphabet{\mathsfit}{bold}{\encodingdefault}{\sfdefault}{bx}{n}

\usepackage{pifont}
\newcommand{\cmark}{\ding{51}}       
\newcommand{\xmark}{\ding{55}}       

\usepackage{array,tabularx,booktabs,multirow,makecell}

\icmltitlerunning{Fire on Motion: Optimizing Video Pass-bands for Efficient Spiking Action Recognition}

\begin{document}

\twocolumn[
  \icmltitle{Fire on Motion: Optimizing Video Pass-bands \\ for Efficient Spiking Action Recognition}



  \icmlsetsymbol{equal}{*}

\begin{icmlauthorlist}
  \icmlauthor{Shuhan Ye}{nbu,ntu,equal}
  \icmlauthor{Yuanbin Qian}{nbu,equal}
  \icmlauthor{Yi Yu}{ntu}
  \icmlauthor{Chong Wang}{nbu}
  \icmlauthor{Yuqi Xie}{nbu}
  \icmlauthor{Jiazhen Xu}{nbu}
  \icmlauthor{Kun Wang}{ntu}
  \icmlauthor{Xudong Jiang}{ntu}
\end{icmlauthorlist}

\icmlaffiliation{nbu}{Ningbo University}
\icmlaffiliation{ntu}{Nanyang Technological University}

\icmlcorrespondingauthor{Chong Wang}{wangchong@nbu.edu.cn}
\icmlcorrespondingauthor{Yi Yu}{yu.yi@ntu.edu.sg}


  \icmlkeywords{Machine Learning, ICML}

  \vskip 0.3in
]



\printAffiliationsAndNotice{\icmlEqualContribution}

\begin{abstract}
Spiking neural networks (SNNs) are adopted in vision tasks for energy efficiency, bio-plausibility, and temporal processing, yet progress remains on static image benchmarks and SNNs lag artificial neural networks (ANNs) on video tasks.
Through analysis, we identify a \emph{pass-band mismatch}. Standard spiking dynamics are known to behave as temporal low-pass filters. While this property has limited impact on static tasks, it can over-emphasize slowly varying components and attenuate motion-related frequency components on video tasks, which often carry critical motion semantics. 
This helps explain why SNNs, despite integrating temporal information, can be constrained on dynamic tasks.
To address this, we propose the Pass-Bands Optimizer (PBO), a plug-and-play module that promotes a task-adaptive temporal pass-band.
PBO uses only two learnable parameters and a lightweight consistency constraint to preserve semantics and boundaries, adds negligible overhead, and requires no architectural changes. 
On UCF101, PBO yields over 10\% improvement. On more challenging multi-modal action recognition and video anomaly detection, it delivers consistent, significant gains, offering a new perspective for SNN-based video processing and understanding.

\end{abstract}

\section{Introduction}
Spiking neural networks (SNNs), the third generation of neural networks \citep{m:97}, have attracted interest for their event-driven computation, biological plausibility, and energy efficiency \citep{snn}. 
Unlike active artificial neural networks (ANNs), SNNs maintain a temporal state that integrates inputs and fires spikes when potentials cross a threshold \citep{LIF}.
This endows SNNs with the capability for temporal processing, yet this machinery remains under-exploited.
Most empirical progress focuses on static tasks (\textit{e.g.,} image classification), creating a pseudo-temporal dimension by replicating a single frame\citep{pfeiffer2018deep}. 
On such benchmarks, recent SNN can match or even exceed strong ANN counterparts while preserving attractive sparsity and low latency \cite{qkformer}. 
Yet when tasks genuinely rely on \emph{temporal} reasoning and motion cues, such as action recognition \citep{rgb1ar, rgb2ar}, video anomaly detection (VAD)~\citep{ucfcrimedvs}, SNNs' performance still falls short of expectations\citep{yu2024svformer,ma2025spiking}.

Empirically, in RGB video streams, abundant static background and low-frequency redundancy elicit large volumes of motion-irrelevant spikes, consuming a limited spiking budget\citep{ASA}. 
Previous work shows that SNNs can benefit from residual inputs relative to RGB \citep{respike}, suggesting that sparse, motion-dominant signals better match spiking computation. 
However, pure frame differencing removes direct current (DC) entirely: it improves sparsity but discards substantial semantic content, making the sparsity-semantics trade-off difficult to measure and optimize. 
Meanwhile, \citet{hfi} show that SNNs tend to attenuate high-frequency content, and mitigate this by boosting \emph{spatial} high frequencies (\textit{e.g.,} via max-pooling) to reduce forward-path degradation. 
However, this doesn't directly address the \emph{temporal} low-pass bias of spiking dynamics, nor does it translate into clear gains on video tasks.

\begin{figure*}[!t]
\centering
\includegraphics[width=\linewidth]{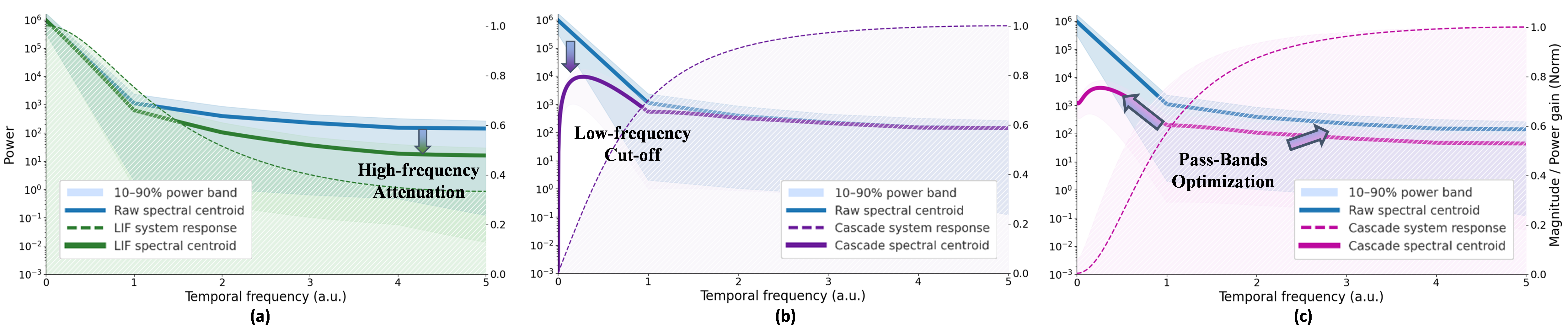}  
\vspace{-6mm}
\caption{Temporal power spectra computed over the full UCF101 dataset \citep{ucf101} and the effects of different filters. 
(a) The LIF dynamics act as a low-pass filter, suppressing high-frequency components. 
(b) Cascading a temporal high-pass with the LIF stage retains high-frequency content but eliminates low-frequency energy, nulling the DC component. 
(c) Our plug-and-play PBO module with LIF adaptively optimizes the temporal pass-band, yielding a learned trade-off between low-frequency content and motion cues.}
\label{fig1}
\vspace{-3mm}
\end{figure*}

To investigate, we revisit SNNs from a frequency-domain perspective and offer a unified diagnosis: a \emph{pass-band mismatch}. 
In SNNs, the LIF neuron's subthreshold membrane potential is equivalent to a linear low-pass transformation of the temporal input \citep{Gerstner2012,hfi}, which preserves content around DC while attenuating the rest. 
Thus, mid-to-high temporal frequencies that vary strongly over time are suppressed to near zero, motion-related semantics are removed.
As shown in Fig.~\ref{fig1} (a), raw videos concentrate energy at DC (blue line). 
Standard SNN processing (green) compounds attenuation at mid-to-high bands and mainly retains DC and low-frequency content. 
In Fig.~\ref{fig1} (b), cascading a first-order temporal high-pass hard-truncates the spectrum at $\omega\!\approx\!0$: it zeros the DC component and strongly attenuates low frequencies, causing a complete loss of low-frequency and steady-state information.

Therefore, raw frames and simple first-order differencing face an unfavorable trade-off between sparsity and semantics.
To resolve this, we introduce the Pass-Band Optimizer (PBO), an extremely simple plug-and-play pre-filter that adaptively modulates membrane integration and shifts the effective temporal response, promoting SNNs to capture task-informative frequency content.
As shown in Fig.~\ref{fig1}(c), with PBO's adaptive pass-band shaping, the learned response retains low-frequency content at a level comparable to mid-to-high motion bands, using only two learnable scalars before embedding with negligible overhead.
Extensive experiments show significant and consistent gains on action recognition and weakly supervised video anomaly detection (VAD), under both \emph{uni-modal} (RGB) and \emph{multi-modal} (RGB + event) settings.
Particularly, it achieves over 10\% improvement on UCF101~\citep{ucf101}. 
PBO is plug-and-play across these tasks, delivering substantial gains with negligible compute overhead.

Our contributions are summarized as below:
\begin{itemize}
\vspace{-2mm}
\item We are the first to diagnose and analyze the \emph{temporal pass-band mismatch} in SNN video processing, providing a frequency-domain perspective for further SNN-based video understanding.
\vspace{-2mm}
\item We propose \emph{Pass-Band Optimizer} (PBO), a plug-and-play causal pre-filter inserted before the membrane integration with only \emph{two} learnable scalars. 
PBO reshapes the response from low-pass to task-aligned pass-band without changing backbones. 
\vspace{-2mm}
\item We introduce a consistency constraint for pass-band optimization and validate our method on multiple motion-centric benchmarks, covering uni-modal (RGB) and multi-modal (RGB+DVS) action recognition and weakly supervised video anomaly detection.
Experimental results demonstrate that PBO achieves stable and significant performance gains, showing robust generalization on dynamic vision tasks.
\end{itemize}

\section{Preliminary and Related Work}\label{sec:rw}
\textbf{Spiking neural networks (SNNs)} replace continuous activation function (\textit{e.g.,} ReLU) with spike neurons, enabling spike-driven sparsity and temporal state via membrane dynamics and resets. 
In this work, we adopt the widely used discrete-time Leaky Integrate-and-Fire (LIF) neuron~\citep{LIF}.
The membrane potential and spike firing of the LIF model are governed by
\begin{equation}\footnotesize
\begin{split}
U[t] = V[t-1] + \tau \left( X[t] - \left( V[t-1] - V_{\text{reset}} \right) \right),\\
S[t] \!=\! \Theta\left( U[t] \!-\! V_{\text{th}} \right),~V[t] \!=\! U[t]  \big(1\!-\!S[t]\big) + V_{\text{reset}}S[t],
\label{eq1}
\end{split}
\end{equation}
where $\tau \triangleq 1/\tau_m$ is the discrete-time leakage rate (inverse membrane time constant), adopted for notational convenience in the subsequent analysis.
$X[t]$ and $V[t]$ denote the input and membrane potential at time step $t$. $U[t]$ is the pre-spike membrane potential while $S[t]\in\{0,1\}^{d}$ is the binary spike computed via the Heaviside function $\Theta(\cdot)$ with firing threshold $V_{\mathrm{th}}$ and the reset potential $V_{\mathrm{reset}}$.

\textbf{SNN-based Discrete-Time Frequency Analysis.}
SNNs possess unique temporal modeling capacity for dynamic vision, and several works have analyzed their frequency characteristics. \citep{fsta} reports strong temporal redundancy across time steps and introduces a frequency-based spatio–temporal attention to suppress it. \citep{hfi} argues SNNs behave as low-pass at the network level and restores missing high-frequency content via extra max-pooling and early depth-wise convolution. However, these approaches remain largely confined to image classification, where task signals are dominated by static appearance. For RGB video understanding, SNN analysis and methods are still scarce; thus we analyze and optimize temporal pass-bands in the frequency domain.

With the above context, we collect the discrete-time frequency preliminaries and notation for our analysis. Given a discrete-time sequence \(x[t]\) of length \(T\), we work directly with its Discrete-Time Fourier Transform (DTFT) of the length-\textit{T} sequence: \(X(e^{j\omega})=\sum_{t=0}^{T-1} x[t]\,e^{-j\omega t}\), where \(j^2=-1\) and \(\omega\in[-\pi,\pi]\) is the normalized angular frequency in radians per sample. Note that \( \omega=0 \) is DC and \( \omega=\pi \) is the Nyquist edge. 
Let \(h[t]\) be an impulse response with frequency response \(H(e^{j\omega})\). 
For a linear time-invariant (LTI) system, its output $y[t]$ is the time-domain convolution of $h[t]$ and $x[t]$, corresponding to the frequency-domain multiplication:
\begin{equation}\footnotesize
\begin{aligned}
y[t]\!=\!h[t]\!*\!x[t]\!=\!\!\!\!\sum_{\tau=-\infty}^{\infty}\!\!\! h[\tau]x[t\!-\!\tau]\!\overset{\mathrm{DTFT}}{\longleftrightarrow}\! Y(e^{j\omega})\!=\!H(e^{j\omega})X(e^{j\omega}).\nonumber
\end{aligned}
\end{equation}
For low-frequency redundancy or contamination, a low-pass stage $H_{\mathrm{LP}}(\cdot,\omega_c)$ with cutoff $\omega_c\!\in(0,\pi)$ preserves baseband and attenuates the high end, with ideal magnitude:
\begin{equation}\footnotesize
\begin{aligned}
\lvert H_{\mathrm{LP}}(e^{j\omega},\omega_c)\rvert=\mathbb{I}\{\,|\omega|\le \omega_c\,\},
\nonumber
\end{aligned}
\end{equation}
where $\mathbb{I}$ is the indicator. Conversely, when high-frequency noise dominates, a high-pass stage $H_{\mathrm{HP}}(\cdot,\omega_c)$ with the same cutoff suppresses DC and retains the high end:
\begin{equation}\footnotesize
\begin{aligned}
\lvert H_{\mathrm{HP}}(e^{j\omega},\omega_c)\rvert=\mathbb{I}\{\,|\omega|\ge \omega_c\,\},
\nonumber
\end{aligned}
\end{equation}
Cascading these two yields a band-pass filter $H_{\mathrm{BP}}(\cdot,\omega_1,\omega_2)$ with $0<\omega_1<\omega_2\le\pi$:
\begin{equation}\footnotesize
\begin{aligned}
H_{\mathrm{BP}}(e^{j\omega},\omega_1,\omega_2)=H_{\mathrm{HP}}(e^{j\omega},\omega_1)\,H_{\mathrm{LP}}(e^{j\omega},\omega_2), \\
\lvert H_{\mathrm{BP}}(e^{j\omega})\rvert=\mathbb{I}\{\,\omega_1\le|\omega|\le\omega_2\,\}.
\nonumber
\end{aligned}
\end{equation}
We then use these transforms and their filter relations, treating membrane integration, leakage and synaptic dynamics as frequency-shaping filters, to analyze dynamic-vision data and learn pass-bands aligned with temporal structure.

\section{Analysis of Pass-bands Mismatch under LIF Constraints}
\label{sec:analysis-sd-lif}
In this work, we model the input to a spiking layer as a discrete-time vector sequence $\bm{X}[t]\in\mathbb{R}^d$ of length $T$, 
which can be decomposed as:
\begin{equation}\footnotesize
\bm{X}[t] \;=\; \bm{B} \;+\; \bm{M}[t] \;+\; \bm{n}[t], \label{eq:decomp}
\end{equation}
where $\bm{B}\in\mathbb{R}^d$ concentrates at DC, $\bm{M}[t]$ captures action-induced dynamics with the angular frequency $\omega>0$, and $\bm{n}[t]$ is additive noise, assuming $\bm{B}$, $\bm{M}[t]$, and $\bm{n}[t]$ are mutually uncorrelated.
At the \emph{subthreshold} steps in a LIF neuron defined by Eq.~\ref{eq1}, the binary spike vector $S[t]=0$, \textit{i.e.,} $V[t]=U[t]$, leading to:
\begin{equation}\footnotesize
\begin{aligned}
U[t] &= V[t-1] \;+\; \tau\!\left(X[t]-\big(V[t-1]-V_{\text{reset}}\big)\right)
\\
&= (1-\tau)\,V[t-1]\;+\;\tau V_{\text{reset}}\;+\;\tau X[t].
\end{aligned}
\end{equation}
Then, recentring $\tilde{\bm{V}}[t]\triangleq \bm{V}[t]-V_{\text{reset}}$ yields the first-order linear recursion:
\begin{equation}\footnotesize
\tilde{\bm{V}}[t] \!=\! \alpha\,\tilde{\bm{V}}[t-1] + (1\!-\!\alpha)\,\bm{X}[t],~\text{where}~
\alpha \triangleq 1-\tau \in (0,1). \label{eq:lif_rec}
\end{equation}
Taking the DTFT over $t$ gives the temporal frequency response, we have:
\begin{equation}\footnotesize
\begin{aligned}
H_{\mathrm{LIF}}(e^{j\omega}) \;=\; \tfrac{\tilde{V}(e^{j\omega})}{X(e^{j\omega})}
\;=\; \tfrac{1-\alpha}{1-\alpha e^{-j\omega}},\\
\text{with} ~
\bigl|H_{\mathrm{LIF}}(e^{j\omega})\bigr|^{2} \;=\;
\tfrac{(1-\alpha)^2}{1+\alpha^{2}-2\alpha\cos\omega}, \label{eq:lif_freq}
\end{aligned}
\end{equation}
which is a classic \textbf{temporal low-pass} with passband near $\omega=0$ and increasing attenuation as $\omega$ grows. 
Let \(S(\omega)\) be the DTFT-based power spectral density (PSD) of a wide-sense stationary input. Since \(X[t]\) admits the linear decomposition in Eq.~\ref{eq:decomp}, the PSD of the LTI input $S_{\mathrm{in}}(\omega)$ can be decomposed into the energies of corresponding components:
\begin{equation}\footnotesize
S_{\mathrm{in}}(\omega)\!\equiv\! S_X(\omega)\!=\!S_B(\omega)\!+\!S_M(\omega)\!+\!S_n(\omega).
\end{equation}
Then, its output $S_{\mathrm{out}}(\omega)$ gives as:
\begin{equation}\footnotesize
\begin{aligned}
S_{\mathrm{out}}(\omega) \;&=\; \bigl|H_{\mathrm{LIF}}(e^{j\omega})\bigr|^{2}\, S_{\mathrm{in}}(\omega)
\;\\
&=\; \bigl|H_{\mathrm{LIF}}(e^{j\omega})\bigr|^{2}\,\Big(S_B(\omega)+S_M(\omega)+S_n(\omega)\Big). \label{eq:lif_psd}
\end{aligned}
\end{equation}
Since $\lvert H_{\mathrm{LIF}}(e^{j0})\rvert^{2}=1$, $S_B(\omega)=S_B(0)$, and $S_M(0)=0$ because motion resides at $\omega>0$, {\textbf{DC}} pass essentially unattenuated while motion is suppressed at nonzero bands:
\begin{equation}\footnotesize
\begin{aligned}
S_{\mathrm{out}}(0) \!&=\! \bigl|H_{\mathrm{LIF}}(e^{j0})\bigr|^{2}\big(S_B(0)\!+\!S_n(0)\big)
\!+\! \bigl|H_{\mathrm{LIF}}(e^{j0})\bigr|^{2} S_M(0) \\
&= S_B(0)+S_n(0). \label{eq:dc_pass}
\end{aligned}
\end{equation}
For any fixed $\omega_0>0$, low-pass attenuation implies:
\begin{equation}\footnotesize
\begin{aligned}
\int_{\omega_0}^{\pi}\! \bigl|H_{\mathrm{LIF}}(e^{j\omega})\bigr|^{2}\,S_M(\omega)\,d\omega
\;\le\; \varepsilon(\alpha,\omega_0)\,\int_{\omega_0}^{\pi}\! S_M(\omega)\,d\omega,\\
\varepsilon(\alpha,\omega_0)\triangleq \max_{\omega\in[\omega_0,\pi]} \bigl|H_{\mathrm{LIF}}(e^{j\omega})\bigr|^{2}\ll 1,
\label{eq:motion_bound}
\end{aligned}
\end{equation}
which formalizes that $\bm{B}$ and low-frequency noise consume spike budget, while the motion-bearing (task-relevant) component $\bm{M}[t]$ is heavily attenuated after the membrane. This explains why SNNs can achieve comparable performance to ANNs on static image tasks, yet struggle on dynamic tasks due to the significant loss of motion information. To be precise, without explicit frequency-domain processing, SNNs primarily rely on the DC and ultra-low-frequency components of the video. We refer to this phenomenon as \textbf{pass-band mismatch}.
To address this issue, we aim to design and cascade a \textbf{learnable pre-filter} $H(e^{j\omega};\theta)$ with the frequency coefficient $\theta$ before the membrane: 
\begin{equation}\footnotesize
S_{\mathrm{out}}^{(\theta)}(\omega) \;=\; \bigl|H(e^{j\omega};\theta)\bigr|^{2}\,\bigl|H_{\mathrm{LIF}}(e^{j\omega})\bigr|^{2}\,S_{\mathrm{in}}(\omega). 
\end{equation}
Let \(\mathcal{G}(\cdot;\theta)\) denotes the model that cascades \(H(e^{j\omega};\theta)\) with the membrane and the task head. Given $N$ pairs of input $X_i[t]$ and label $y_i$, and supervised loss \(\ell\) (\textit{e.g.,} cross-entropy), the optimization is:
\begin{equation}\footnotesize
\min_{\theta}\ \mathcal{L}(\theta)=\frac{1}{N}\sum_{i=1}^{N}\ell(\mathcal{G}(X_i[t];\theta), y_i).
\label{eq:theta_sup}
\end{equation}

\section{Methodology}
\label{sec:method-bpa}
\textbf{Motivation.}
Despite the temporal modeling capacity of LIF neurons, our analysis (Sec.~\ref{sec:analysis-sd-lif}) shows that the LIF constraint induces a temporal low-pass whose pass-band is \textbf{mismatched} with video dynamics. 
We therefore insert a \emph{learnable, causal} pre-filter \emph{before} the embedding stack and optimize it during training to obtain a task-informative pass-band. In addition, we introduce a \emph{consistent loss} to find a dynamic balance between low-pass and high-pass systems.

\subsection{Temporal Pre-filter and Cascaded Response}
\label{subsec:prefilter}
\textbf{Definition.}
We first define the pass-bands pre-filter with a two-point shift-and-subtract (with $\lambda\!\in\![0,1]$):
\begin{equation}\footnotesize
\bm{Y}^{\!(\lambda)}\![t] \!=\! \bm{X}[t] \!-\! \lambda\bm{X}[t\!-\!1]
\!=\! (1\!-\lambda)\bm{X}[t] \!+\! \lambda\big(\bm{X}[t]\!-\!\bm{X}[t\!-\!1]\big).
\label{eq:pre}
\end{equation}
This operation can be expressed as a learnable weighted sum of the original frame and the frame difference, making it minimal and lightweight. Its frequency response is:
\begin{equation}\footnotesize
\begin{aligned}
W(e^{j\omega},\lambda)=1-\lambda e^{-j\omega},\\
\text{with}~ \bigl|W(e^{j\omega},\lambda)\bigr|^{2}=1+\lambda^{2}-2\lambda\cos\omega.
\label{eq:pbo-freq}
\end{aligned}
\end{equation}

\textbf{Cascaded Frequency Response.}
Using the LIF temporal frequency response in Eq.~\ref{eq:lif_freq}, the cascaded transfer becomes:
\begin{equation}\footnotesize
\begin{aligned}
\!\!G(e^{j\omega},\lambda)
\!=\! W(e^{j\omega},\lambda)\,H_{\mathrm{LIF}}(e^{j\omega})\!=\! \frac{(1\!-\!\lambda e^{\!-\!j\omega})(1\!-\!\alpha)}{1-\alpha e^{-j\omega}},\\
\bigl|G(e^{j\omega},\lambda)\bigr|^{2}
\!=\!  \frac{(1\!+\!\lambda^{2}\!-\!2\lambda\cos\omega)\,(1\!-\!\alpha)^{2}}{\,1+\alpha^{2}-2\alpha\cos\omega\,}.
\end{aligned}
\label{eq:cascade-mag}
\end{equation}
However, with fixed \(\alpha\), Eq.~\ref{eq:cascade-mag} implies that as \(\lambda\) sweeps \(0\!\to\!1\), the DC gain \(|G(e^{j0})|^{2}=(1-\lambda)^{2}\) decreases while the high-frequency endpoint \(|G(e^{j\pi})|^{2}=(1+\lambda)^{2}(1-\alpha)^{2}{(1+\alpha)^{-2}}\) increases. The response is monotone in \(\omega\), \textit{i.e.,} low-pass tilt if \(\lambda<\alpha\), flat at \(\lambda=\alpha\) and high-pass tilt if \(\lambda>\alpha\). Thus, a single \(\lambda\) only shifts the passband centroid and cannot form a mid-band peak or independently control bandwidth (more details in Appendix~\ref{subsec:passband-single-lambda}). We therefore generalize to a \textbf{time-varying} \(\lambda[t]\).

\vspace{-1mm}
\subsection{Pass-bands Optimizer}
\vspace{-2mm}
\textbf{LTV pass-bands optimizer definition.}
To enlarge the optimizable pass-band in both shape and spectral center, we generalize the scalar \(\lambda\) to a \textbf{time-varying sequence} \(\lambda[t]\), yielding a two-tap linear time-varying (LTV) pre-filter. 
We use a single-layer instantiation throughout this work.
The formulation in Eq.~\ref{eq:pre} is thus redefined:

\vspace{-4mm}
\begin{equation}\footnotesize
\begin{aligned}
\bm{Y}[t] \;=\; \bm{X}[t] \;-\; \lambda[t]\bm{X}[t-1],\\ \!\!\text{where}~
h[t,0]\!=\!1,~h[t,1]\!=\!-\lambda[t],~h[t,k]\!=\!0\ (k\!\notin\!\{0,1\}).
\label{eq:LTV}
\end{aligned}
\end{equation}

\textbf{A plug-and-play stationarity-aware periodic pre-filter.}
To maintain architectural simplicity and support seamless deployment, such LTV pass-bands optimizer is designed as a plug-and-play module to generate a time-varying coefficient sequence \(\lambda[t]\) before the first spiking membrane, without modifying the backbone or inference flow. 
To respect spectral stationarity and enable interpretable frequency-domain shaping, we adopt a \emph{bounded-energy, mean-stable, cyclostationary} parameterization of \(\lambda[t]\) determined by two learnable parameters $\mu$ and $\omega$, which preserves a well-defined DC baseline with controllable frequency-sideband structures. Concretely, we set:
\begin{equation}\footnotesize
\begin{aligned}
\lambda[t] \;=\; \mu \;+\; A\,\sin(\omega\,t+\phi),\\ 
\text{where}~
\mu\in[0,1],~A\ge 0,~ \omega\in(0,\pi],~ \phi\in\mathbb{R}.
\label{eq:param-lambda}
\end{aligned}
\end{equation}
\(\mu\) determines the time-average (DC component) of \(\lambda[t]\), governing the mean behavior of the pre-filter (\textit{e.g.,} stronger DC suppression as \(\mu\) increases). Meanwhile, the sinusoidal modulation introduces structured nonzero-frequency components that broaden and shape the effective pass-band.

\textbf{Parameterization and initialization.}
Given the learnable mean \(\mu\in[0,1]\) and angular frequency \(\omega \in (0,\pi)\), \(\omega\) is indirectly determined by a learnable raw variable \(\sigma_{\mathrm{raw}}\in\mathbb{R}\) via a logistic map:
$p = \sigma(\sigma_{\mathrm{raw}}) = \frac{1}{1+e^{-\sigma_{\mathrm{raw}}}},~\omega = \pi\,p.$
%
The mean \(\mu\) determines the DC baseline and averages pass-band tilt, initialized at \(\mu=0.5\). It is then guided toward a dynamic equilibrium within \([0,1]\) via a consistency loss defined later in Eq.~\ref{eq:consist}. For default initialization over a clip of length \(T\), we target one period via \(\omega_0 = \tfrac{2\pi}{T-1}\) and set
$\sigma_{\mathrm{raw}} \;\leftarrow\; \log\tfrac{2/(T-1)}{1 - 2/(T-1)}.$

\textbf{Harmonic-transfer view and dominant sidebands.}
Under a \(P\)-periodic \(\lambda[t]\) with fundamental \(\omega_0=2\pi/P\), the Linear Periodically Time-Varying system (LPTV) response admits the harmonic-transfer representation:
\begin{equation}\footnotesize
Y(e^{j\omega}) \;=\; \sum_{m\in\mathbb{Z}} W_m(e^{j\omega})\,X\!\big(e^{j(\omega-m\omega_0)}\big),
\label{eq:LPTV}
\end{equation}
where \(\{W_m\}\) are determined by the Fourier coefficients of \(\lambda[t]\). For the single-tone model in Eq.~\ref{eq:param-lambda} with \(\phi=0\) and \(\omega=\omega_0\), we have:
\begin{equation}\footnotesize
\begin{aligned}
\!\!\lambda_0\!=\!\mu,
\lambda_{\pm1}\!=\!\mp\frac{A}{2j},
\lambda_{|m|>1}\!=\!0\;{\Rightarrow}W_0(e^{j\omega}) \!=\! 1\!-\!\mu\,e^{\!-\!j\omega},\\
W_{\pm1}(e^{j\omega}) \!=\! \!-\!\,\lambda_{\pm1}\,e^{\!-\!j\omega},~
W_{|m|>1}(e^{j\omega})\!=\!0.
\end{aligned}
\end{equation}

\textbf{Cascade with LIF and PSD approximation.}
By the response in Eq.~\ref{eq:lif_freq}, the cascaded spectrum is:
\begin{equation}\footnotesize
Y(e^{j\omega})
\;=\;
H_{\mathrm{LIF}}(e^{j\omega})
\sum_{m\in\mathbb{Z}} W_{m}(e^{j\omega})\,X\!\big(e^{j(\omega-m\omega_{0})}\big).
\label{eq:cascade-ltv}
\end{equation}
{For convenience of derivation, we assume approximate uncorrelatedness across frequency bins (the full correlated version is derived in Appendix~\ref{correlated}, and the resulting sideband conclusion remains unchanged)}.
The output PSD is approximated by:
\begin{equation}\footnotesize
\begin{aligned}
S_{\mathrm{out}}(\omega)\ \approx\
\bigl|H_{\mathrm{LIF}}(e^{j\omega})\bigr|^{2}\!
\Big(
\bigl|W_{0}(e^{j\omega})\bigr|^{2}\,S_{\mathrm{in}}(\omega)
\\
+\sum_{m\neq 0}\bigl|W_{m}(e^{j\omega})\bigr|^{2}\,S_{\mathrm{in}}(\omega-m\omega_{0})
\Big).
\end{aligned}
\label{eq:PSD-out}
\end{equation}
For the single-tone case \(\omega=\omega_0\) with \(\phi=0\), it can be simplified as:
\begin{equation}\footnotesize
\begin{aligned}
S_{\mathrm{out}}(\omega)\ \approx\
\bigl|H_{\mathrm{LIF}}(e^{j\omega})\bigr|^{2}\!
\Big(
\underbrace{\bigl|1-\mu\,e^{-j\omega}\bigr|^{2} S_{\mathrm{in}}(\omega)}_{\text{baseline (DC) term}}
\\
+\underbrace{\tfrac{A^{2}}{4}\,S_{\mathrm{in}}(\omega-\omega_{0})+\tfrac{A^{2}}{4}\,S_{\mathrm{in}}(\omega+\omega_{0})}_{\text{frequency-translation sidebands}}
\Big),
\end{aligned}
\label{eq:psd-simplified}
\end{equation}
which reveals that the time variation injects controllable sidebands at \(\pm\omega_0\) which \textbf{translate low-frequency energy into nonzero bands}. Choosing \(\omega_0\) inside the LIF-transmissible region yields a genuine mid-band pass window, while \(\mu\) sets the DC floor and hyper-parameter \(A\) controls peak height and effective bandwidth.

\textbf{Why this is reasonable despite time variation?}
Even though \(\lambda[t]\) is time-varying, our design remains theoretically sound and practically stable for the following reasons:
\textbf{(i)} The time-average $\frac{1}{T}\sum_{t=1}^{T}\lambda[t]\in[0,1]$ preserves the physical interpretability of the two-tap pre-emphasis filter, while the bounded magnitude of $\lambda[t]$ supports numerical stability in practice.
\textbf{(ii)} From an expectation perspective, the time-varying coefficient sequence behaves equivalently to a constant filter with \(\lambda = \mu\), thereby retaining the original high-pass tilt and DC suppression characteristics of the static formulation. Detailed derivation is in Appendix~\ref{ltv}.
\textbf{(iii)} Since \(\lambda[t]\) is a structured periodic function, it induces a cyclostationary filtering regime that legitimizes the use of harmonic-transfer analysis in Eq.~\ref{eq:LPTV}--\ref{eq:psd-simplified}, providing a mechanism to broaden and shape the effective passband while remaining streaming and computationally lightweight.

\textbf{Consistency loss.}
\label{consistent-loss}
To keep the optimized pre-LIF signal faithful to the original semantics, we regularize the learned mixture against the two endpoints in Eq.~\ref{eq:pre}. Let \(\bm{Y}^{(0)}[t]=\bm{X}[t]\) (DC) and \(\bm{Y}^{(1)}[t]=\bm{X}[t]-\bm{X}[t-1]\) (High-frequency), and denote the filtered output by \(\bm{Y}^{(m)}_t\). A weight \(\lambda[t]\) balances the two references. 
To avoid bias toward either endpoint caused by absolute intensity scale, we \emph{spatially de-mean} the signals in the intensity term. Let
\begin{equation}\footnotesize
\tilde{\bm{Y}}^{(k)}_t = \bm{Y}^{(k)}_t - \frac{1}{HW}\sum_{x=1}^{W}\sum_{y=1}^{H}\bm{Y}^{(k)}_t(x,y),~k\in\{0,1,m\},
\end{equation}
computed per time step and per channel. The consistency loss can be formulated as:
\begin{equation}\footnotesize
\begin{aligned}
\mathcal{L}_{\mathrm{consist}}
&=\frac{1}{TCHW}\sum_{t=0}^{T-1}\big(\mathcal{L}^{\mathrm{int}}_{t}+\mathcal{L}^{\mathrm{grad}}_{t}\big),\\
\mathcal{L}_{t}^{\mathrm{int}}
&=\left\| \lambda[t] \odot \big(\tilde{\bm{Y}}^{(m)}_{t}-\tilde{\bm{Y}}^{(1)}_{t}\big) \right\|_2^2
\\
&+\left\| (1-\lambda[t]) \odot\big(\tilde{\bm{Y}}^{(m)}_{t}-\tilde{\bm{Y}}^{(0)}_{t}\big) \right\|_2^2,\\
\mathcal{L}^{\mathrm{grad}}_{t}
&=\left\| \nabla_x \bm{Y}^{(m)}_t - \max(|\nabla_x \bm{Y}^{(0)}_t|,|\nabla_x \bm{Y}^{(1)}_t|) \right\|_1
\\
&+\left\| \nabla_y \bm{Y}^{(m)}_t - \max(|\nabla_y \bm{Y}^{(0)}_t|,|\nabla_y \bm{Y}^{(1)}_t|) \right\|_1
\end{aligned}
\label{eq:consist}
\end{equation}
where \(\nabla_x,\nabla_y\) are Sobel gradients along horizontal and vertical directions. The \(L_2\) term penalizes per-pixel deviations from the \emph{demeaned} endpoints according to \(\lambda[t]\), thereby preventing intensity-scale attraction to either side and yielding a balanced objective. The \(L_1\) term aligns edges by matching the stronger response from either endpoint, preserving sharpness and structural sparsity.

\textbf{Closed-form equilibrium of the intensity term.}
For fixed \(\lambda[t]\), minimizing \(\mathcal{L}^{\mathrm{int}}_{t}\) over \(\tilde{\bm{Y}}^{(m)}_{t}\) admits a per-pixel closed form:
\begin{equation}\footnotesize
\begin{aligned}
\hat{\tilde{\bm{Y}}}^{\!(m)}_{t}
&\!=\!\arg\!\min_{\tilde{\bm{Y}}}\!\! \left[\lambda[t]^2\!\left\|\tilde{\bm{Y}}\!-\!\tilde{\bm{Y}}^{(1)}_{t}\right\|_2^2
\!+\!\big(1\!-\!\lambda[t]\big)^2\!\left\|\tilde{\bm{Y}}\!-\!\tilde{\bm{Y}}^{(0)}_{t}\right\|_2^2
\right]\\
&\!=\!\frac{\lambda[t]^2\,\tilde{\bm{Y}}^{(1)}_{t}+\big(1-\lambda[t]\big)^2\,\tilde{\bm{Y}}^{(0)}_{t}}{\lambda[t]^2+\big(1-\lambda[t]\big)^2},
\end{aligned}\nonumber
\end{equation}
which interpolates between the low-pass (\(\lambda{=}0, \hat{\tilde{\bm{Y}}}^{\!(m)}_{t}\!\!=\!\!\tilde{\bm{Y}}^{\!(0)}_{t}\)) and the high-pass (\(\lambda{=}1, \hat{\tilde{\bm{Y}}}^{\!(m)}_{t}\!\!=\!\!\tilde{\bm{Y}}^{\!(1)}_{t}\)), establishing a well-posed dynamic equilibrium within \([0,1]\).
Overall, we train the model and the pass-bands optimizer with a classification loss and two auxiliary terms $\mathcal{L}
= \mathcal{L}_{\mathrm{ce}}
+ \alpha\;\mathcal{L}_{\mathrm{consist}}$.

\begin{table*}[t]
\centering
\footnotesize
\setlength{\tabcolsep}{12.5pt}
\renewcommand{\arraystretch}{1.0}
\caption{Results of different backbones with vs.\ without PBO on UCF101 and HMDB51.}
\label{tab:main_results}
\vspace{-2mm}
\scalebox{0.95}{%
\begin{tabular}{c c c c c c c}
\toprule
\textbf{Dataset} & \textbf{Methods} & \textbf{Architecture} & \textbf{Params} & $T$ & \textbf{Acc(\%)} & \textbf{$\Delta$(\%)} \\
\midrule

\multirowcell{4}{\rotatebox{90}{\textbf{UCF101}}}
& Spikformer~\citep{zhou2023spikformer} & Spikformer-2-256 & 2.58M & 10 & 46.16$^{\ast}$ & -- \\
\cmidrule{2-7}
& \textbf{Spikformer + PBO} & \textbf{Spikformer-2-256}
    & \textbf{2.58M} & \textbf{10} & \textbf{57.71} & \textbf{+11.55} \\
\cmidrule{2-7}
& SDT-V1~\citep{sd-transformer} & SD-Transformer-2-256 & 2.59M & 10 & 49.25$^{\ast}$ & -- \\
\cmidrule{2-7}
& \textbf{SDT-V1 + PBO} & \textbf{SD-Transformer-2-256}
    & \textbf{2.59M} & \textbf{10} & \textbf{59.80} & \textbf{+10.55} \\
\midrule

\multirowcell{4}{\rotatebox{90}{\textbf{HMDB51}}}
& Spikformer~\citep{zhou2023spikformer} & Spikformer-2-256 & 2.58M & 10 & 58.66$^{\ast}$ & -- \\
\cmidrule{2-7}
& \textbf{Spikformer + PBO} & \textbf{Spikformer-2-256}
    & \textbf{2.58M} & \textbf{10} & \textbf{65.22} & \textbf{+6.56} \\
\cmidrule{2-7}
& SDT-V1~\citep{sd-transformer} & SD-Transformer-2-256 & 2.59M & 10 & 62.24$^{\ast}$ & -- \\
\cmidrule{2-7}
& \textbf{SDT-V1 + PBO} & \textbf{SD-Transformer-2-256}
    & \textbf{2.59M} & \textbf{10} & \textbf{68.21} & \textbf{+5.97} \\
\bottomrule
\end{tabular}%
}
\vspace{-3mm}
\end{table*}

\begin{figure*}[h]
\centering
\includegraphics[width=\linewidth]{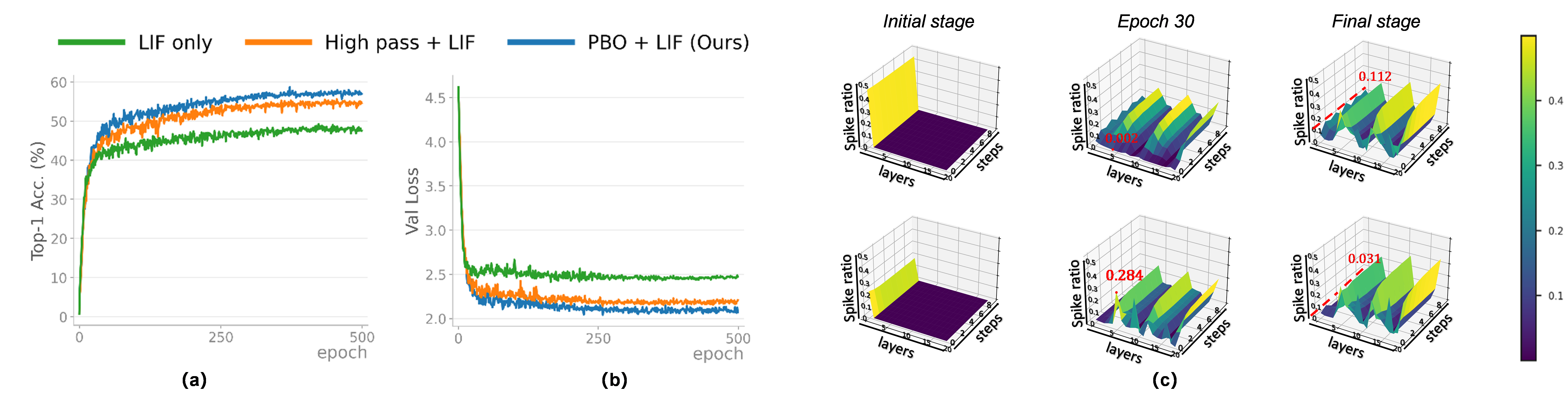}  
\vspace{-6mm}
\caption{UCF101 results with a spike-driven transformer \citep{sd-transformer}.
(a) Top-1 accuracy vs.~epoch for three schemes: LIF only (low-pass), High-pass $\rightarrow$ LIF (coarse band-pass), and PBO $\rightarrow$ LIF (ours).
(b) Corresponding validation loss.
(c) Layer-step spike-ratio surfaces of the LIF across training. Left$\rightarrow$right: initial, epoch~30, final. Top row: LIF only, bottom row: PBO$\rightarrow$LIF.
Color encodes spike ratio in $[0,0.5]$. \emph{Red dots indicate the firing ratio used for query mapping}.}
\label{fig3}
\vspace{-3mm}
\end{figure*}

\section{Experiments}
\label{sec:ex}
\textbf{Datasets and Tasks.}
We evaluate four kinds of dynamic vision tasks: uni-modal action recognition, multi-modal action recognition, uni-modal and multi-modal video anomaly detection. All experiments use \emph{color and event paired} (CEP) datasets in which RGB and dynamic vision sensor (DVS) streams are spatio-temporally aligned. 
Since DVS reports per-pixel intensity changes asynchronously with microsecond temporal resolution and sparse activity, the corresponding stream is motion dominant.
\begin{enumerate}
\vspace{-2mm}
\item \textbf{UCF101-CEP.}
UCF101~\citep{ucf101} has 13{,}320 RGB videos over 101 classes. Its DVS counterpart, UCF101-DVS~\citep{ucf101dvs}, is generated with the DAVIS240 simulator. The public DVS release contains 13{,}523 clips, which exceeds the RGB set. Thus, we remove redundancies to enforce one-to-one pairing. Since about half of the DVS clips are horizontally flipped relative to RGB, we flip RGB ones to match the orientation. We refer to the aligned pair set as \textbf{UCF101-CEP}.
\vspace{-2mm}
\item \textbf{HMDB51-CEP.}
HMDB51~\citep{hmdb} has 6{,}766 RGB clips in 51 classes. 
Paired with its DVS counterpart~\citep{ucf101dvs} generated with the DAVIS240, referred as \textbf{HMDB51-CEP}.
\vspace{-2mm}
\item \textbf{HARDVS.}
HARDVS~\citep{hardvs} is the largest DVS action recognition dataset recorded with DAVIS346, containing $>$100{,}000 clips over 300 classes. It offers naturally captured, temporally aligned RGB-DVS streams and is challenging due to scale, diversity, and realistic conditions.
\vspace{-2mm}
\item \textbf{UCF-Crime-CEP.}
UCF-Crime and UCF-Crime-DVS~\citep{ucfcrimedvs} form the largest VAD set with RGB and event modalities. We use both inputs and compare against representative ANN and SNN baselines on this fine-grained anomaly detection task.
\end{enumerate}

\vspace{-2mm}
\textbf{Implementation Details.}
Experiments are conducted on the BrainCog platform~\citep{braincog} using four NVIDIA RTX 4090 GPUs. We train all models with AdamW (initial learning rate 0.005). Unless stated otherwise, the LIF node uses a time constant $\tau=0.7$ and a firing threshold of 1, {the amplitude $A$ and the phase $\phi$ are set to 0.1 and 0 on all the datasets}. To assess the plug-and-play nature of PBO, we compare against baselines taken from the strongest numbers reported in the original papers when available. Otherwise, we reimplement the methods under their stated settings and use the reproduced accuracy as the baseline. We then insert PBO into the same backbones without changing the architecture, loss, data preprocessing, training schedule, or compute budget, and we report improvements under the same sequence length and input resolution.

\begin{table*}[t]
\centering
\footnotesize
\setlength{\tabcolsep}{10.0pt}
\renewcommand{\arraystretch}{1.0}
\caption{Comparison with existing methods on UCF101-DVS, HMDB51-DVS, and HARDVS datasets. \textbf{$^{\ast}$} Results reproduced under our unified implementation framework.}
\vspace{-2mm}
\scalebox{0.95}{
\begin{tabular}{c c c c c c c}
\toprule
\textbf{Dataset} & \textbf{Category} & \textbf{Methods} & \textbf{Architecture} & \textbf{Params} & $T$ & \textbf{Accuracy} \\
\midrule
\multirow{12}{*}{\rotatebox{90}{\textbf{UCF101-DVS}}}
& \multirow{3}{*}{ANN}
& 3D CNN~\citep{c3d}              & C3D                     & 78.41M  & 8/16 & 38.2 / 47.2 \\
& & RG-CNN~\citep{ucf101dvs}      & RG-CNN + Incep. 3D      & 6.95M   & 8/16 & 63.2 / 67.8 \\
& & ESCNet~\citep{ECSNet}         & ESCNet-SES              & --      & 8/16 & 59.9 / 70.2 \\
\cmidrule{2-7}
& \multirow{3}{*}{SNN}
& RM-SNN~\citep{RMSNN}            & ResNet-18               & --      & 8    & 58.5 \\
& & TIM~\citep{tim}               & Spikformer-2-256        & 2.58M   & 10   & 63.8 \\
& & TIM~\citep{tim}               & SD-Transformer-2-256    & 2.59M   & 10   & 64.38$^{\ast}$ \\
\cmidrule{2-7}
& \multirow{6}{*}{\shortstack{Multi-modal\\SNN}}
& SCA~\citep{SCA}                 & Spikformer-2-256        & 3.60M   & 10   & 60.11$^{\ast}$ \\
& &WeiAttn~\citep{WAt}             & SD-Transformer-2-256    & 3.33M   & 10   & 67.58$^{\ast}$ \\
& &CMCI~\citep{CMCI}               & SD-Transformer-2-256    & 4.44M   & 10   & 65.69$^{\ast}$ \\
& &S-CMRL~\citep{avspikformer}     & SD-Transformer-2-256    & 4.10M   & 10   & 68.13$^{\ast}$ \\
& &\textbf{S-CMRL + PBO (Ours)}           & \textbf{SD-Transformer-2-256} & \textbf{4.10M}  & \textbf{10} & \textbf{73.03} \\

\midrule
\multirow{12}{*}{\rotatebox{90}{\textbf{HMDB51-DVS}}}
& \multirow{3}{*}{ANN}
& 3D CNN~\citep{c3d}              & C3D                     & 78.41M  & 8/16 & 34.2 / 41.7 \\
& & RG-CNN~\citep{ucf101dvs}      & RG-CNN + Incep. 3D      & 6.95M   & 8/16 & 45.2 / 51.5 \\
& & I3D~\citep{I3D}               & I3D                     & 12.37M  & 8/16 & 38.6 / 46.6 \\
\cmidrule{2-7}
& \multirow{3}{*}{SNN}
& RM-SNN~\citep{RMSNN}            & ResNet-18               & --      & 8    & 44.7 \\
& & TIM~\citep{tim}               & Spikformer-2-256        & 2.58M   & 10   & 58.6 \\
& & TIM~\citep{tim}               & SD-Transformer-2-256    & 2.59M   & 10   & 61.93$^{\ast}$ \\
\cmidrule{2-7}
& \multirow{6}{*}{\shortstack{Multi-modal\\SNN}}
& SCA~\citep{SCA}                 & Spikformer-2-256        & 3.60M   & 10   & 70.15$^{\ast}$ \\
& &WeiAttn~\citep{WAt}             & SD-Transformer-2-256    & 3.32M   & 10   & 71.94$^{\ast}$ \\
& &CMCI~\citep{CMCI}               & SD-Transformer-2-256    & 4.40M   & 10   & 71.64$^{\ast}$ \\
& &S-CMRL~\citep{avspikformer}     & SD-Transformer-2-256    & 4.09M   & 10   & 72.33$^{\ast}$ \\
& &\textbf{S-CMRL + PBO (Ours)}           & \textbf{SD-Transformer-2-256} & \textbf{4.09M}  & \textbf{10} & \textbf{74.18} \\
\bottomrule
\end{tabular}
}
\vspace{-3mm}
\label{tab:cep_results}
\end{table*}

\begin{table*}[!t]
\caption{Comparison of accuracy and energy consumption on HARDVS dataset.}
\vspace{-2mm}
\label{tab:model_energy_comparison}
\centering
\footnotesize
\renewcommand{\arraystretch}{1.0}
\setlength{\tabcolsep}{9.0pt}
\scalebox{0.95}{
\begin{tabular}{c c c c c c c c}
\toprule
\textbf{Type} & \textbf{Method} & \textbf{Model} & \textbf{Params} & \textbf{$T$} & \textbf{Energy (mJ)} & \textbf{$\Delta$ (\%)} & \textbf{Acc (\%)} \\
\midrule
\multirow{4}{*}{ANN} 
    & ACTION-Net~\citep{action-net}   & ResNet-50  & 27.9M   & 8 & --& --  & 46.85 \\
    &TimeSformer~\citep{timesformer} & ViT-B/16   & 121.2M  & 8  & --& -- & 50.77 \\
    & ESTF \citep{hardvs} & ResNet-18       & 46.7M   & 8   & 81.1  & -- & 51.2 \\
    & TSM \citep{tsm}     & ResNet-50       & --   & 8   & 87.4  & -- & 52.6 \\
\midrule
\multirow{3}{*}{SNN} 
    & SDT-V1 \citep{sd-transformer} & SDT-V1  & 2.6M   & 8   & --    & -- & 36.5 \\
    & SDT-V2 \citep{sdv2}           & SDT-V2 & 18.3M   & 8   & 8.0   & -- & 47.5 \\
    & SDT-V3 \citep{sdv3}           & SDT-V3    & 18.7M   & 8   & 23.5  & -- & 49.2 \\
\midrule
\multirow{8}{*}{\makecell[c]{Multi- modal \\ SNN}}
    & WeiAttn \citep{WAt}       & SDT-V1 & 3.36M & 8 & 0.145 & -- & 48.6 \\
    & \textbf{WeiAttn + PBO}    & \textbf{SDT-V1} & \textbf{3.36M} & \textbf{8} & \textbf{0.133} & \textbf{$\downarrow$ 8.3} & \textbf{49.1} \\
    \cmidrule(lr){2-8}
    & SCA \citep{SCA}           & SDT-V1 & 3.65M & 8 & 0.162 & -- & 48.8 \\
    & \textbf{SCA + PBO}        & \textbf{SDT-V1} & \textbf{3.65M} & \textbf{8} & \textbf{0.142} & \textbf{$\downarrow$ 12.3} & \textbf{49.7} \\
    \cmidrule(lr){2-8}
    & CMCI \citep{CMCI}         & SDT-V1 &  4.42M & 8 & 0.174 & -- & 48.1 \\
    & \textbf{CMCI + PBO}       & \textbf{SDT-V1} & \textbf{4.42M} & \textbf{8} & \textbf{0.155} & \textbf{$\downarrow$ 10.9} & \textbf{49.2} \\
    \cmidrule(lr){2-8}
    & S-CRML \citep{avspikformer}   & SDT-V1 & 4.16M & 8 & 0.168 & -- & 49.7 \\
    & \textbf{S-CRML + PBO}         & \textbf{SDT-V1} & \textbf{4.16M} & \textbf{8} & \textbf{0.147} & \textbf{$\downarrow$ 12.5} & \textbf{51.3} \\
\bottomrule
\end{tabular}
}
\vspace{-2mm}
\end{table*}

\subsection{Main Experiments}
\textbf{Uni-modal effectiveness on RGB video datasets.}
As shown in Table~\ref{tab:main_results}, our plug-and-play Pass-Band Optimizer (PBO) brings large gains on UCF101 and HMDB51. On UCF101, PBO lifts Spikformer from \textbf{46.16\% to 57.71\%} and SDT-V1 from \textbf{49.25\% to 59.80\%}. On HMDB51, we observe similar consistent improvements: \textbf{58.66\% to 65.22\%} for Spikformer and \textbf{62.24\% to 68.21\%}for SDT-V1. These gains arrive without altering the backbone, underscoring that aligning a model’s temporal pass-band with task-informative motion bands is a first-order factor for RGB-based action recognition in SNNs. More backbone results with PBO are reported in Table~\ref{tab:backbone} of the appendix.

From the learning dynamics presented in Fig.~\ref{fig3} (a) and (b), PBO reaches \textit{lower} validation loss than either (i) a coarse band-pass formed by naively cascading High-pass $\rightarrow$ LIF, or (ii) the LIF-only low-pass. While convergence speed in epochs is comparable, the final loss plateau of our PBO is markedly smaller, indicating a better optimal solution rather than a mere optimization acceleration. 

The spike-ratio surfaces given in Fig.~\ref{fig3} (c) also support our motivation and theory. By \textbf{epoch 30}, PBO has already activated the $Q_{\text{LIF}}$ pathway within the spiking self-attention (SSA) \citep{spikingformer}, producing structured layer-step selective firing (particularly in mid-temporal steps), whereas the LIF-only baseline remains largely quiescent and under-responsive. 
By the final stage, PBO maintains sparse yet \textit{functionally engaged} activity patterns, consistent with a well-shaped band-pass. Collectively, these results show that PBO not only improves accuracy substantially but also steers the network toward a more semantically aligned and energetically disciplined operating regime.

\textbf{Multi-modal effectiveness with DVS.}
Beyond uni-modal action recognition, we evaluate RGB-DVS fusion to further test its effectiveness. PBO is utilized as a drop-in module in the RGB branch, so that it adaptively optimizes the temporal pass-band, which improves complementarity with the DVS stream.
Under a unified implementation, attaching PBO to the spiking fusion method S-CMRL achieves \textbf{73.03\%}, \textbf{74.18\%}, \textbf{51.30\%} accuracy on UCF101-CEP, HMDB51-CEP and HARDVS, respectively, as shown in Table~\ref{tab:cep_results} and Table~\ref{tab:model_energy_comparison}. This corresponds to gains of \textbf{+4.90}, \textbf{+1.85} and \textbf{+1.60} percentage points over the baseline without PBO (S-CMRL). It can be seen that the proposed PBO surpasses recent uni-modal methods (TIM, SDT-V2, SDT-V3) and multi-modal SNN fusion methods (WeiAttn, SCA, CMCI). 
These gains come without modifying backbone, indicating that simple plug-and-play pass-band alignment is sufficient to unlock strong uni-modal and multi-modal improvements. 

\subsection{Discussion}
\vspace{-2mm}

\begin{table}[t]
\centering
\caption{Ablation on UCF101-CEP for Leaky factor $\tau$, Consistency weight $\alpha$ ($\times 10^{-3}$), and Amplitude $A$ in $\lambda[t]=\mu + A\sin(\omega t+\phi)$.}
\vspace{-2mm}
\label{tab:ablation_all}
\footnotesize
\renewcommand{\arraystretch}{1.0}
\setlength{\tabcolsep}{6pt}
\scalebox{0.90}{
\begin{tabular}{l|cccccc}
\toprule
$\tau$   & 0.3 & 0.5 & 0.7 & 0.9 \\
Accuracy (\%) & 71.27 & 72.57 & \textbf{73.03} & 70.20 \\
\midrule
$\alpha$ ($10^{-3}$)   & 0 & 1 & 5 & 10 & 30 & 50 \\
Accuracy (\%) & 64.70 & 70.41 & 72.49 & \textbf{73.03} & 72.41 & 70.78 \\
\midrule
$A$ & 0 & 0.1 & 0.3 & 0.5 \\
Accuracy (\%) & 70.14 & \textbf{73.03} & 72.17 & 72.25 \\
\bottomrule
\end{tabular}}
\vspace{-2mm}
\end{table}

\begin{table}[t]
\caption{Ablation on $\mathcal{L}_{\mathrm{consist}}$ components on UCF101-CEP.}
\vspace{-2mm}
\label{tab:loss}
\centering
\footnotesize
\renewcommand{\arraystretch}{1.0}
\setlength{\tabcolsep}{10pt}
\scalebox{0.90}{
\begin{tabular}{c c c}
\toprule
$\mathcal{L}_{t}^{\mathrm{int}}$ & $\mathcal{L}_{t}^{\mathrm{grad}}$ & Accuracy (\%) \\
\midrule
\xmark & \xmark & 64.70 \\
\cmark & \xmark & 72.17 \\
\xmark & \cmark & 69.64 \\
\cmark & \cmark & \textbf{73.03} \\
\bottomrule
\end{tabular}
}
\vspace{-6mm}
\end{table}

\noindent\textbf{Ablation studies.}
As shown in Table~\ref{tab:ablation_all}, we ablate the leak factor $\tau$, the consistency weight $\alpha$, and the modulation amplitude $A$.
\textbf{1) Leak factor $\tau$: }On UCF101-CEP, the leak factor $\tau$ exhibits a clear ``middle is best" trend: accuracies at $\tau\!=\!0.3/0.5/0.7/0.9$ are $71.27\%/72.57\%/\mathbf{73.03}\%/70.20\%$.
Within the $\lambda[t]$ based system, $\tau\!=\!0.7$ induces a relatively strong leaky low-pass that cooperates best with our PBO.
Interestingly, across different $\tau$, the learned $\mu$ consistently converges near $1$, while the learned $\omega$ varies substantially. Such frequency changes with $\tau$ highlights the necessity of pass-band shifting.
Detailed visualizations are provided in Appendix~\ref{sec:vis-mu-omega}.
\textbf{2) Consistency weight $\alpha$:}
Removing the term ($\alpha\!=\!0$) reduces accuracy to $64.70\%$, indicating that without the consistency regularizer the optimized pre LIF signal may drift from the original semantics and introduce distortion.
Increasing $\alpha$ to $1{\times}10^{-3}$ and $5{\times}10^{-3}$ improves accuracy to $70.41\%$ and $72.49\%$. It peaks at $\alpha\!=\!\mathbf{1{\times}10^{-2}}$ with $\mathbf{73.03}\%$, while larger values ($3{\times}10^{-2}$, $5{\times}10^{-2}$) begin to hurt.
\textbf{3) Modulation amplitude $A$:}
A small $A\!=\!0.1$ already reaches the highest accuracy $73.03\%$.
Increasing $A$ to $0.3$ and $0.5$ yields slight drops but remains stable overall.
Removing this term ($A\!=\!0$) degenerates the original time varying system into a time invariant one. As a result, the pass-band cannot be modulated and the accuracy drops sharply to $70.14\%$, further validating  our theory and the necessity of pass-band shifting.
The ablation on HMDB51-CEP shows the similar trend, which can be found in Table \ref{tab:hyper_new} of the supplementary materials.
\textbf{4) Consistency loss components: }
Table~\ref{tab:loss} ablates the proposed consistency loss. Removing it causes a clear drop (64.70\%), indicating that unconstrained pass-band optimization can drift away from semantically meaningful inputs. Using only $\mathcal{L}_{t}^{\mathrm{int}}$ already yields a large gain (72.17\%), while $\mathcal{L}_{t}^{\mathrm{grad}}$ alone also improves performance (69.64\%) by preserving edge and structural cues. Combining both terms achieves the best result (73.03\%), suggesting that intensity fidelity and boundary preservation are complementary and jointly stabilize pass-band learning.
To further validate the robustness of PBO, we report more ablation studies under different clip lengths, sampling strides, and input resolutions in Tables~\ref{tab:clip_length}, \ref{tab:stride}, and \ref{tab:resolution}, respectively, in the Appendix \ref{app:more-exp}.


\textbf{Energy Evaluation.}
We also compare the energy consumption and recognition accuracy of representative ANN, SNN, and multimodal SNN methods on HARDVS, which are summarized in Table~\ref{tab:model_energy_comparison}.
Our method attains a favorable balance between performance and efficiency.
The measurement protocol and computation details are provided in Appendix~\ref{energy-evaluation}.
Compared with uni-modal and multi-modal SNN baselines, PBO reaches $51.3\%$ accuracy with only $0.147\,\mathrm{mJ}$, surpassing all SNN based methods. It further reduces energy while improving accuracy over all multi-modal fusion methods.

\begin{table}[t]
\centering
\caption{
Results on UCF-Crime and UCF-Crime-DVS.}
\vspace{-2mm}
\label{tab:VADMain}
\footnotesize
\renewcommand{\arraystretch}{1.0}
\setlength{\tabcolsep}{1pt}
\scalebox{0.9}{
\begin{tabular}{ccccc}
\toprule
\textbf{Type} & \textbf{Method} & \textbf{Features} & \textbf{AUC(\%)} & \textbf{FAR(\%)} \\
\midrule
\multirow{4}{*}{ANN}
& \citet{sultani2018real} & Event & 55.56 & 8.69 \\
& 3C-Net~\citep{narayan20193c}          & Event & 59.22 & 9.50 \\
& AR-Net~\citep{wan2020weakly}          & Event & 60.71 & 8.51 \\
& RTFM~\citep{tian2021weakly}           & Event & 52.67 & 13.19 \\
\midrule
\multirow{5}{*}{SNN}
& \multirow{3}{*}{MSF~\citep{ucfcrimedvs}} & Event        & 65.01 & \textbf{3.27} \\
&                                          & RGB          & 71.54 & 14.54 \\
&                                          & RGB + Event  & 70.01 & 17.89 \\
\cmidrule{2-5}
& \multirow{2}{*}{MSF + PBO}               & RGB                   & 72.31 & 10.89 \\
&                                          & \textbf{RGB + Event}  & \textbf{74.14} & 5.19 \\
\bottomrule
\end{tabular}
}
\vspace{-6mm}
\end{table}

\textbf{Extension to Video Anomaly Detection.}
The VAD task requires frame level anomaly scoring over long videos with sparse and irregular events, which makes conventional ANN pipelines energy intensive.
To assess scalability under weak supervision, we apply our plug-and-play PBO on the MSF \citep{ucfcrimedvs}, implementation details and visualizations are provided in Appendix~\ref{vad}.
As shown in Table~\ref{tab:VADMain}, our PBO can effectively improve RGB only VAD method by increasing AUC (Area Under the ROC Curve) and reducing FAR (False Alarm Rate).
In the RGB$+$DVS setting, MSF combined with PBO achieves state-of-the-art performance.

\vspace{-4mm}
\section{Conclusion}
\vspace{-2mm}
This paper reframes SNN video processing as a temporal \emph{pass-band mismatch} and shows that SNNs face a fundamental trade-off between achieving suitable sparsity and preserving rich semantics on video tasks.
We introduce Pass-Band Optimizer (PBO), a plug-and-play, causal pre-filter that reshapes the LIF-induced low-pass adaptively toward a task-informative pass-band. 
PBO adds only two lightweight scalars and requires no backbone changes. 
Empirically, PBO delivers consistent gains across dynamic tasks on UCF101-CEP, HMDB51-CEP, and HARDVS, and extends to weakly supervised VAD on UCF-Crime-CEP, achieving favorable accuracy and energy under the same training budgets. We believe this frequency-oriented view opens avenues for SNN-based video understanding.

\newpage

\section*{Impact Statement}
This work proposes a plug-and-play pass-band optimizer for spiking neural networks (SNNs) for video action recognition and weakly supervised video anomaly detection. We evaluate only on public benchmarks with paired RGB and event (DVS) streams—UCF101/UCF101-DVS (aligned as UCF101-CEP), HMDB51/HMDB51-DVS (HMDB51-CEP), HARDVS, and UCF-Crime/UCF-Crime-DVS; no new data were collected, and no human subjects or personally identifiable information are involved. Aware of dual-use risks in video understanding (\textit{e.g.,} surveillance), we restrict our study to public datasets, release any artifacts for research-only use, provide no deployment-oriented functionality, and do not target identification or re-identification. For reproducibility and environmental responsibility, we report implementation details, keep compute modest (training on four NVIDIA RTX~4090 GPUs), and estimate energy using standard SNN synaptic-operation accounting with established CMOS energy costs, reporting accuracy--energy trade-offs and encouraging license-compliant, responsible use.


\bibliography{main}
\bibliographystyle{icml2026}

\newpage
\appendix
\onecolumn
\section{Pass-band Characteristics and the Limitation of a Single \texorpdfstring{$\lambda$}{lambda}}
\label{subsec:passband-single-lambda}

Using the cascaded magnitude response in Eq.~\ref{eq:cascade-mag}, and for readability letting
\(a=1+\lambda^{2}\), \(b=2\lambda\), \(c=1+\alpha^{2}\), \(d=2\alpha\),
we can rewrite the pass-band as
\begin{equation}
\bigl|G(e^{j\omega},\lambda)\bigr|^{2}
\;=\;(1-\alpha)^{2}\,\frac{a-b\cos\omega}{\,c-d\cos\omega\,}.
\label{eq:passband-fraction}
\end{equation}

\paragraph{Endpoint gains (delimiting the band edges).}
\begin{equation}
\bigl|G(e^{j0},\lambda)\bigr|^{2}=(1-\lambda)^{2},
\qquad
\bigl|G(e^{j\pi},\lambda)\bigr|^{2}
=\frac{(1+\lambda)^{2}(1-\alpha)^{2}}{(1+\alpha)^{2}}.
\label{eq:endpoints}
\end{equation}

\paragraph{Tilt vs. flat point (no mid-band peak with a single \(\lambda\)).}
Differentiating \(\frac{a-b\cos\omega}{c-d\cos\omega}\) w.r.t. \(u=\cos\omega\) yields
\[
\frac{d}{du}\!\left(\frac{a-bu}{c-du}\right)
=\frac{ad-bc}{(c-du)^{2}},
\qquad
ad-bc=2(\alpha-\lambda)(1-\alpha\lambda).
\]
Hence, for fixed \(\alpha\):
\begin{equation}
\begin{split}
\lambda<\alpha:\ \text{peak at }\omega=0\ \text{(low-pass tilt)};\\ \quad 
\lambda=\alpha:\ \bigl|G\bigr|^{2}\equiv(1-\alpha)^{2}\ \text{(flat)};\quad \\
\lambda>\alpha:\ \text{peak at }\omega=\pi\ \text{(high-pass tilt)}.
\end{split}
\label{eq:tilt-flat}
\end{equation}
\textit{Implication.} A single scalar \(\lambda\) can only move the passband centroid from low to high frequencies (with a flat point at \(\lambda=\alpha\)); it cannot create a genuine mid-band peak, \textit{i.e.,} a strict band-pass window.

\paragraph{\(-3\) dB cutoffs (unique solutions).}
Because \(|G|^{2}\) is strictly monotone over \(\omega\in[0,\pi]\) when \(\lambda\neq\alpha\), each tilt has a unique \(-3\) dB cutoff.

\textbf{Low-pass tilt} \((\lambda<\alpha)\), normalized at \(\omega=0\):
\begin{equation}
\frac{1+\lambda^{2}-2\lambda\cos\omega_{c}^{(\mathrm{LP})}}
     {1+\alpha^{2}-2\alpha\cos\omega_{c}^{(\mathrm{LP})}}
=\frac{1}{2}\cdot\frac{(1-\lambda)^{2}}{(1-\alpha)^{2}}.
\label{eq:cut-lp}
\end{equation}

\textbf{High-pass tilt} \((\lambda>\alpha)\), normalized at \(\omega=\pi\):
\begin{equation}
\frac{1+\lambda^{2}-2\lambda\cos\omega_{c}^{(\mathrm{HP})}}
     {1+\alpha^{2}-2\alpha\cos\omega_{c}^{(\mathrm{HP})}}
=\frac{1}{2}\cdot\frac{(1+\lambda)^{2}}{(1+\alpha)^{2}}.
\label{eq:cut-hp}
\end{equation}

Solving Eq.~\ref{eq:cut-lp} or Eq.~\ref{eq:cut-hp} for \(\cos\omega_{c}\in[-1,1]\) gives the unique cutoff frequency. In practice, \(\lambda\) thus controls the passband tilt and edge location under a fixed \(\alpha\), but cannot introduce a mid-band bump without extending to a time-varying \(\lambda[t]\).

\paragraph{Limitation of a single \(\lambda\).}
A scalar \(\lambda\) can only \emph{interpolate} the passband centroid between a low-pass tilt, the flat point, and a high-pass tilt.
It \emph{cannot} create a truly \emph{peaked mid-band} (a strict band-pass window), hence the tunable passband shape and center are limited.

\section{Energy Evaluation}
\label{energy-evaluation}
Energy consumption is a critical metric for evaluating the performance of SNNs. We estimate the theoretical energy consumption following the methodology in~\citep{sd-transformer}. First, the number of synaptic operations (SOPs), which reflect the total number of accumulate (AC) operations triggered by spikes, are estimated as:
\begin{equation}
\text{SOP}_l = R_l \times T \times \text{FLOP}_l,
\label{eq:sop}
\end{equation}
where $R_l \in [0, 1]$ denotes the average spike rate in layer $l$, $T$ is the number of timesteps, and $\text{FLOP}_l$ is the number of floating-point operations in the corresponding non-spiking layer. On 45\,nm CMOS hardware, the energy cost per multiply-accumulate (MAC) operation is $E_{\text{MAC}} = 4.6$\,pJ, while the cost per accumulate (AC) operation is $E_{\text{AC}} = 0.9$\,pJ. The total energy consumption is computed as:
\begin{equation}
E_{\text{total}} = E_{\text{MAC}} \times \text{FLOP}_1 + E_{\text{AC}} \times \sum_{l=2}^{L} \text{SOP}_l,
\label{eq:energy}
\end{equation}
where $\text{FLOP}_1$ denotes the number of floating-point operations in the first convolutional layer. For all subsequent layers ($l \geq 2$), spike-driven binary activations are used, and computations are modeled as synaptic operations $\text{SOP}_l$.

\noindent\textbf{PBO overhead.}
Using the equivalent form $\tilde X[t]=X[t]-\lambda[t]\,X[t-1]$, PBO incurs only \textbf{one multiplication and one addition/subtraction} per element per time step, i.e., $O(THWC)$ operations. Compared to the network forward whose first spiking embedding/patch embedding typically costs $O\!\left(THW\cdot k^{2} C C_{\text{out}}\right)$ (and subsequent blocks are even heavier), this overhead is negligible in practice.

\section{VAD Implementation Details}
\label{vad}
Following \citep{sultani2018real,ucfcrimedvs}, each video and event stream are divided into 16 non-overlapped clips and the total number of training epochs is set to 20.
\subsubsection{Evaluation Metrics} Following prior works \citep{wan2020weakly,ucfcrimedvs}, we report Area Under of Curve (AUC) of the frame-level Receiver Operating Characteristics (ROC) and False Alarm Rate (FAR) with a threshold 0.5. AUC measures the overall discriminative capability of the model, while FAR evaluates its reliability and robustness in real-world scenarios.
\subsubsection{Experiment Results} As shown in Table~\ref{tab:VADMain}, our method alone achieves an accuracy of 65.45\%, which increases to 71.41\% when combined with MSF, outperforming the previous MSF by 0.44\% and 6.40\%, respectively. 
This demonstrates that PBO not only sets a new performance benchmark for SNNs in weakly supervised video anomaly detection, but also that the integration with MSF highlights the effectiveness of our PBO in providing temporally and semantically coherent features that are more amenable to SNN learning.
These results validate our method both as a feature enhancement module and a viable backbone, marking a significant step toward closing the performance gap between SNNs and ANNs in this domain.

\subsection{Visualization} 
\begin{figure*}[!t]
\centering
\includegraphics[width=\textwidth]{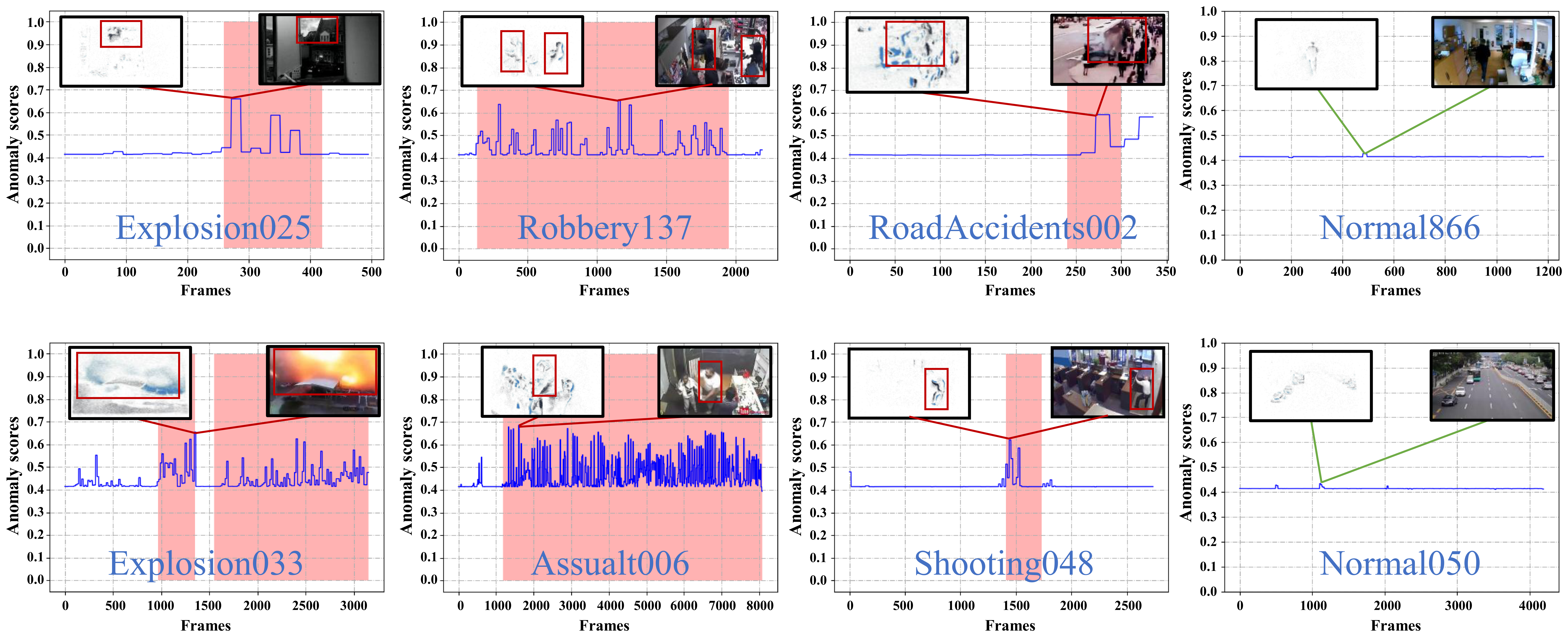}  
\vspace{-6mm}
\caption{Anomaly scores of our methods on Color-Event UCF-Crime. Pink areas indicate the manually labelled abnormal events, purple lines represent the anomaly score and red boxes point out abnormal events on the screen.}
\label{fig7}
\end{figure*}

Fig.~\ref{fig7} presents a set of visualizations demonstrating that PBO effectively distinguishes normal from abnormal events. For instance, in Shooting048, the anomaly scores rise sharply when individuals raise and fire guns. In Robbery137, although the anomaly scores do not consistently exceed the threshold throughout the anomalous segments, this is attributed to relatively static scenes that fail to trigger event responses in the DVS, resulting in partial information loss. Nevertheless, elevated scores are observed during key moments, such as gun possession and the act of stealing from cabinets. For explosion events, which share visual patterns with scene transitions or light flickering in DVS data, PBO is able to differentiate them accurately, thereby reducing false alarms.

\section{Why LTV is reasonable?}
\label{ltv}
\paragraph{Approximation to a constant-\(\lambda\) filter.}
Let \(\lambda[t]=\mu+\delta[t]\) with time average \(\frac{1}{T}\sum_{t=0}^{T-1}\delta[t]\to 0\) and variance \(\frac{1}{T}\sum_{t=0}^{T-1}\delta[t]^2\to \sigma_\lambda^2\).
The instantaneous frequency response of the two-tap pre-filter is
\begin{equation}\footnotesize
H_t(e^{j\omega}) \;=\; 1 - \big(\mu+\delta[t]\big)e^{-j\omega}.
\end{equation}
Averaging the squared gain over time yields
\begin{equation}\footnotesize
\overline{|H|^2}(\omega)
\;\triangleq\;
\lim_{T\to\infty}\frac{1}{T}\sum_{t=0}^{T-1}\!\big|H_t(e^{j\omega})\big|^2
=
\big|1-\mu e^{-j\omega}\big|^2
+\lim_{T\to\infty}\frac{1}{T}\sum_{t=0}^{T-1}\delta[t]^2,
\end{equation}
because the cross-term vanishes by the zero-mean assumption on \(\delta[t]\). Hence
\begin{equation}\footnotesize
\overline{|H|^2}(\omega)
\;=\;
\underbrace{\big|1-\mu e^{-j\omega}\big|^2}_{\text{constant-\(\lambda\) template}}
\;+\;\sigma_\lambda^2
\;=\;
\big(1+\mu^2-2\mu\cos\omega\big)\;+\;\sigma_\lambda^2.
\label{eq:avg-gain}
\end{equation}
Therefore, when the variance \(\sigma_\lambda^2\) is small (or treated as an \(\omega\)-independent offset), the time-varying filter \(\lambda[t]\) is well approximated by the constant-\(\lambda\) filter with \(\lambda=\mu\) in the sense of average squared gain.

\paragraph{High-pass property preserved at the mean.}
The constant-\(\lambda\) template satisfies
\begin{equation}\footnotesize
\big|1-\mu e^{-j\omega}\big|^2
= 1+\mu^2-2\mu\cos\omega
\;\approx\; (1-\mu)^2 + \mu\,\omega^2 \quad (\omega\to 0),
\end{equation}
so \(\omega=0\) is a local minimum for any \(\mu>0\), \textit{i.e.,} the response is high-pass around DC. Since \(\sigma_\lambda^2\) in Eq.~\ref{eq:avg-gain} is \(\omega\)-independent, the high-pass shape (low-frequency suppression) is preserved under the approximation \(\lambda[t]\approx \mu\).

\section{Visualization of $\mu$ and $\omega$ in $\lambda[t]$}
\label{sec:vis-mu-omega}
\begin{figure*}[h]
\centering
\includegraphics[width=\textwidth]{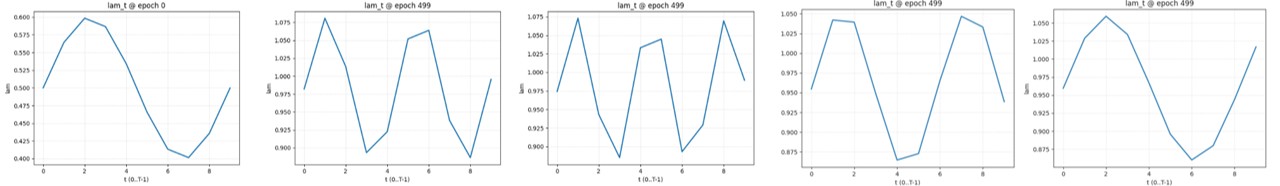}  
\caption{\textbf{Visualization of the learned temporal mixing weights} $\lambda[t]$ under different leak factors $\tau$.
From left to right: initialization (epoch 0), and results at epoch 499 for $\tau\!=\!0.3,\,0.5,\,0.7,\,0.9$.}
\label{vis-mu-omega}
\vspace{-10pt}  
\end{figure*}

\noindent\textbf{Analysis of the modulation patterns in Fig.~\ref{vis-mu-omega}.}
Figure~\ref{vis-mu-omega} plots the learned temporal mixing weights $\lambda[t]$ over $T{=}10$ steps. The leftmost panel shows the initialization (epoch 0), which is a gentle single-period sinusoid. After training (epoch 499), the four panels correspond to $\tau\!=\!0.3,\,0.5,\,0.7,\,0.9$, respectively. As $\tau$ increases, $\lambda[t]$ exhibits progressively higher temporal frequency—roughly from $\sim\!1$ period at initialization to $\gtrsim\!2$ periods when $\tau$ is large—while remaining centered near $\mu\!\approx\!1$ with a small amplitude ($A\!\approx\!0.05$--$0.1$). This behavior is consistent with the LIF update
\[
U[t]\;=\;(1-\tau)\,V[t-1]\;+\;\tau\,X[t]\;+\text{const},
\]
which yields a first-order low-pass whose memory shortens as $\tau$ grows. To complement this shorter memory and the wider pass-band of the LIF cell, the modulator increases its switching rate (higher $\omega$), i.e., it mixes the DC and differential references more rapidly within the same window. 

\section{Full PSD Derivation Without the Uncorrelated-Frequency Assumption}
\label{correlated}

This section provides the complete derivation of the output power spectral density (PSD)
when frequency components of the input signal may be correlated.
No decorrelation assumption is used here.
The result shows that the translated sidebands induced by temporal modulation
remain present regardless of the correlation structure.

\subsection{Spectrum of the cascaded LIF--modulation system}
From Eq.~\eqref{eq:lif_freq}, the output spectrum under a general harmonic modulation
is
\begin{equation}\footnotesize
Y(e^{j\omega})
=
H_{\mathrm{LIF}}(e^{j\omega})
\sum_{m\in\mathbb{Z}}
W_{m}(e^{j\omega})
X\!\left(e^{j(\omega-m\omega_{0})}\right).
\label{eq:appendix-cascade}
\end{equation}
The corresponding PSD is
\begin{equation}\footnotesize
S_{\mathrm{out}}(\omega)
=
\mathbb{E}\!\left[\,\bigl|Y(e^{j\omega})\bigr|^{2}\right].
\end{equation}

\subsection{Expansion of the PSD}
Substituting Eq.~\eqref{eq:appendix-cascade} into the above yields
\begin{equation}\footnotesize
\begin{aligned}
S_{\mathrm{out}}(\omega)
&=
\bigl|H_{\mathrm{LIF}}(e^{j\omega})\bigr|^{2}
\sum_{m}\sum_{n}
W_{m}(e^{j\omega})W_{n}^{*}(e^{j\omega})\,
\mathbb{E}\!\Big[
X\!\left(e^{j(\omega-m\omega_{0})}\right)
X^{*}\!\left(e^{j(\omega-n\omega_{0})}\right)
\Big].
\end{aligned}
\label{eq:appendix-start}
\end{equation}

Define the generalized cross-spectrum
\begin{equation}\footnotesize
S_{X}(\alpha,\beta)
\;=\;
\mathbb{E}\!\left[
X(e^{j\alpha})\,X^{*}(e^{j\beta})
\right],
\end{equation}
then Eq.~\eqref{eq:appendix-start} becomes
\begin{equation}\footnotesize
S_{\mathrm{out}}(\omega)
=
\bigl|H_{\mathrm{LIF}}(e^{j\omega})\bigr|^{2}
\sum_{m}\sum_{n}
W_{m}(e^{j\omega})W_{n}^{*}(e^{j\omega})
S_{X}\!\left(\omega-m\omega_{0},\ \omega-n\omega_{0}\right).
\label{eq:appendix-fullPSD}
\end{equation}
This is the full PSD without approximations.

\subsection{Decomposition into diagonal and cross-spectral components}
The terms with $m=n$ correspond to the auto-spectral contributions:
\begin{equation}\footnotesize
S_{X}(\omega-m\omega_{0},\,\omega-m\omega_{0})
=
S_{\mathrm{in}}(\omega-m\omega_{0}).
\end{equation}
The remaining terms with $m\neq n$ collect all cross-spectral correlations:
\begin{equation}\footnotesize
S_{X}(\omega-m\omega_{0},\,\omega-n\omega_{0}),\qquad m\neq n.
\end{equation}

Thus, the PSD may be written as
\begin{equation}\footnotesize
\begin{aligned}
S_{\mathrm{out}}(\omega)
&=
\bigl|H_{\mathrm{LIF}}(e^{j\omega})\bigr|^{2}
\Bigg[
\underbrace{
\sum_{m}
\bigl|W_{m}(e^{j\omega})\bigr|^{2}\,
S_{\mathrm{in}}(\omega-m\omega_{0})
}_{\text{auto-spectral terms}}
\\[1mm]
&\quad+
\underbrace{
\sum_{m\neq n}
W_{m}(e^{j\omega})W_{n}^{*}(e^{j\omega})\,
S_{X}\!\left(\omega-m\omega_{0},\,\omega-n\omega_{0}\right)
}_{\text{cross-spectral terms}}
\Bigg].
\end{aligned}
\label{eq:appendix-split}
\end{equation}

\subsection{Single-tone modulation case}
For the single-tone modulation used in the main paper, only $m\in\{0,\pm1\}$ are non-zero.
Let
\[
X_{0}=X(\omega),\quad
X_{+}=X(\omega-\omega_{0}),\quad
X_{-}=X(\omega+\omega_{0}),
\]
and similarly for $W_{0},W_{+},W_{-}$.
Substituting into Eq.~\eqref{eq:appendix-split} yields
\begin{equation}\footnotesize
\begin{aligned}
S_{\mathrm{out}}(\omega)
&=
\bigl|H_{\mathrm{LIF}}(e^{j\omega})\bigr|^{2}
\Big[
|W_{0}|^{2} S_{\mathrm{in}}(\omega)
+|W_{+}|^{2} S_{\mathrm{in}}(\omega-\omega_{0})
+|W_{-}|^{2} S_{\mathrm{in}}(\omega+\omega_{0})
\\[1mm]
&\quad
+2\Re\!\Big(
W_{0}W_{+}^{*}\,S_{X}(\omega,\omega-\omega_{0})
+
W_{0}W_{-}^{*}\,S_{X}(\omega,\omega+\omega_{0})
+
W_{+}W_{-}^{*}\,S_{X}(\omega-\omega_{0},\omega+\omega_{0})
\Big)
\Big].
\end{aligned}
\label{eq:appendix-single-tone}
\end{equation}

\subsection{Implication for the learned pass-band}
The terms
\[
|W_{+}|^{2} S_{\mathrm{in}}(\omega-\omega_{0}),
\qquad
|W_{-}|^{2} S_{\mathrm{in}}(\omega+\omega_{0}),
\]
correspond to the frequency-translated sidebands characteristic of harmonic modulation.
These terms remain present irrespective of the cross-spectral correlations
in the input signal. The correlation-dependent expressions
\[
S_{X}(\omega,\omega\!-\!\omega_{0}),\quad
S_{X}(\omega,\omega\!+\!\omega_{0}),\quad
S_{X}(\omega\!-\!\omega_{0},\omega\!+\!\omega_{0})
\]
modify amplitudes but cannot cancel the translated components.
Consequently, the learned modulation continues to produce
a nonzero mid-band emphasis even in the fully correlated case.

\begin{table}[!t]
\caption{Ablation on clip length $T$ on UCF101-CEP (stride = 1).}
\label{tab:clip_length}
\centering
\footnotesize
\renewcommand{\arraystretch}{1.0}
\setlength{\tabcolsep}{7pt}
\scalebox{0.90}{
\begin{tabular}{c c c c}
\toprule
$T$ & Stride & PBO & Top-1 Acc (\%) \\
\midrule
4  & 1 & w/o & 49.33 \\
4  & 1 & w/  & 56.67 \\
\midrule
8  & 1 & w/o & 66.65 \\
8  & 1 & w/  & 72.60 \\
\midrule
10 & 1 & w/o & 68.13 \\
10 & 1 & w/  & 73.03 \\
\midrule
16 & 1 & w/o & 59.95 \\
16 & 1 & w/  & 67.53 \\
\bottomrule
\end{tabular}
}
\vspace{-2mm}
\end{table}

\begin{table}[!t]
\caption{Ablation on temporal sampling stride on UCF101-CEP ($T=8$).}
\label{tab:stride}
\centering
\footnotesize
\renewcommand{\arraystretch}{1.0}
\setlength{\tabcolsep}{7pt}
\scalebox{0.90}{
\begin{tabular}{c c c c}
\toprule
$T$ & Stride & PBO & Top-1 Acc (\%) \\
\midrule
8 & 1 & w/o & 66.65 \\
8 & 1 & w/  & 72.60 \\
\midrule
8 & 2 & w/o & 66.22 \\
8 & 2 & w/  & 71.26 \\
\midrule
8 & 4 & w/o & 65.90 \\
8 & 4 & w/  & 70.81 \\
\bottomrule
\end{tabular}
}
\vspace{-2mm}
\end{table}

\begin{table}[!t]
\caption{Ablation on input spatial resolution on UCF101-CEP ($T=10$, stride = 1).}
\label{tab:resolution}
\centering
\footnotesize
\renewcommand{\arraystretch}{1.0}
\setlength{\tabcolsep}{7pt}
\scalebox{0.90}{
\begin{tabular}{c c c}
\toprule
Resolution  & PBO & Top-1 Acc (\%) \\
\midrule
32  & w/o & 59.68 \\
32  & w/  & 67.74 \\
\midrule
64  & w/o & 68.13 \\
64  & w/  & 73.03 \\
\midrule
128 & w/o & 65.82 \\
128 & w/  & 72.87 \\
\bottomrule
\end{tabular}
}
\vspace{-2mm}
\end{table}


\begin{table}[!t]
\vspace{-2mm}
\centering
\renewcommand{\arraystretch}{1.0}
\caption{Ablation on HMDB-CEP for leaky factor $\tau$, consistency weight $\alpha$, and modulation amplitude $A$ in $\lambda[t]=\mu + A\sin(\omega t+\phi)$. We report Top-1 accuracy (\%).}
\label{tab:hyper_new}
\footnotesize
\setlength{\tabcolsep}{2.5pt}
\scalebox{1.0}{
\begin{tabular}{c|cccc|cccccc|cccc}
\toprule
\multirow{2}{*}{\shortstack{Module\\$\rightarrow$}}
& \multicolumn{4}{c|}{Leaky factor $\tau$}
& \multicolumn{6}{c|}{Consistency weight $\alpha$ ($10^{-3}$)}
& \multicolumn{4}{c}{Amplitude $A$}
\\
& 0.3 & 0.5 & 0.7 & 0.9
& 0 & 1 & 5 & 10 & 30 & 50
& 0 & 0.1 & 0.3 & 0.5
\\
\midrule
Acc (\%) &
73.43 & 73.58 & \textbf{74.18} & 73.13
& 71.94 & 72.99 & 73.43 & \textbf{74.18} & 73.58 & 72.24
& 71.64 & \textbf{74.18} & 74.18 & 72.98
\\
\bottomrule
\end{tabular}
}
\vspace{-4mm}
\end{table}

\begin{table}[h]
\vspace{-2mm}
\centering
\renewcommand{\arraystretch}{1.0}
\caption{Effect of PBO on additional SNN backbones (UCF101-CEP).}
\label{tab:backbone}
\footnotesize
\setlength{\tabcolsep}{5pt}
\scalebox{1.0}{
\begin{tabular}{c|c|c|c|c}
\toprule
Backbone & PBO & Top-1 Acc (\%) & $\Delta$ (\%) & Comment \\
\midrule
QKFormer4-384 & w/o & 45.20 & -- & Baseline \\
QKFormer4-384 & w/  & \textbf{54.62} & +9.42 & Clear improvement \\
\midrule
SVFormer-base & w/o & 63.55 & -- & Baseline \\
SVFormer-base & w/  & \textbf{69.37} & +5.82 & Strong gain \\
\bottomrule
\end{tabular}
}
\vspace{-4mm}
\end{table}

\section{More Experiments}
\label{app:more-exp}

This section provides additional experimental results referenced in the main paper, together with extended ablations for a more complete analysis of the proposed PBO module. We report results on: (1) clip length, (2) temporal sampling stride, (3) input spatial resolution, (4) hyperparameter sensitivity of the leaky factor $\tau$, consistency weight $\alpha$, and modulation amplitude $A$, (5) additional evaluations on two modern SNN backbones (QKFormer and SVFormer). Unless otherwise specified, all experiments follow the same training protocol and implementation details as in the main paper.

\paragraph{Limitations.}
This paper focuses on establishing PBO as a simple, modular component and clarifying its spectral role in spiking video recognition. 
As a result, we intentionally keep the design space narrow and the presentation mechanism-oriented, leaving several orthogonal aspects as future directions, such as alternative parameterizations of the temporal modulator, complexer
more expressive insertion patterns (e.g., multi-stage compositions or depth-dependent placement of PBO).
In addition, devising supervision signals that generalize well across datasets for optimizing PBO remains an interesting open direction.
We believe these extensions are largely complementary to the current contributions rather than required for them.


\end{document}